\documentclass[letterpaper, 10 pt, conference]{ieeeconf}  % Comment this line out if you need a4paper
\usepackage[T1]{fontenc}
\usepackage{times}
\usepackage{amsmath,amssymb}
\usepackage{amsfonts}
\usepackage{tikz-cd}

\usepackage{adjustbox}
\usepackage{array}
\usepackage{subcaption}
\usepackage{booktabs}
\usepackage{multirow}
\usepackage{wrapfig}
\usepackage{tcolorbox}

\IEEEoverridecommandlockouts                              % This command is only needed if 
\title{\LARGE \bf
RMPs for Safe Impedance Control in Contact-Rich Manipulation
}

\author{Seiji Shaw, Ben Abbatematteo, and George Konidaris\\
\texttt{ \{sshaw4, babbatem,gdk\}@cs.brown.edu}% <-this % stops a space
\thanks{Department of Computer Science, Brown University, Providence RI}% <-this % stops a space
}

\begin{document}

\maketitle
\thispagestyle{empty}
\pagestyle{empty}

%%%%%%%%%%%%%%%%%%%%%%%%%%%%%%%%%%%%%%%%%%%%%%%%%%%%%%%%%%%%%%%%%%%%%%%%%%%%%%%%
\begin{abstract}
Variable impedance control in operation-space is a promising approach to learning contact-rich manipulation behaviors. One of the main challenges with this approach is producing a manipulation behavior that ensures the safety of the arm and the environment. Such behavior is typically implemented via a reward function that penalizes unsafe actions (e.g. obstacle collision, joint limit extension), but that approach is not always effective and does not result in behaviors that can be reused in slightly different environments. We show how to combine Riemannian Motion Policies, a class of policies that dynamically generate motion in the presence of safety and collision constraints,  with variable impedance operation-space control to learn safer contact-rich manipulation behaviors.
\end{abstract}

\section{Introduction}

% Paragraph 1: Importance of variable impedance operation-space control for contact-rich tasks
% \begin{itemize}
%     \item Learning contact-rich behavior is hard - mode change, unknown dynamics, happens in a %space removed from joints of a robot
%     \item Martin-Martin showed that SE(3) + Stiffness in SE(3) is an excellent action space
%     \item Allows agent to reason about pose in actual space where the task happens
%     \item Impedance control gives agent finer control for contact-rich tasks, while still %reasoning about free-space, which is fundamentally easier
%     \item Since we only reason about the end-effector, learning can be transferred from robot %to robot
% \end{itemize}

Learning autonomous contact-rich manipulation behavior is a critical challenge for robotics; indeed, the very purpose of  a robot is often to  make contact with the environment in order to manipulate it. However, contact-rich behavior is challenging---robots must be able to reason about sudden constraints when contacting target objects and model the unknown dynamics of the objects they are manipulating, all the while respecting joint limit and collision constraints. %Moreover, many robot arms designed for human collaboration  have more than six degrees of freedom, which  introduces the additional challenge of reasoning about over-actuated systems. 

To address these issues,  Martin-Martin et al. \cite{2019MartinMartinVICES} discuss the advantages of formulating the task in the robot's end-effector space. They propose the use of variable impedance control in end-effector space (VICES), which defines the robot's actions in terms of displacements and compliance of the end-effector. The VICES commands are then translated back as lower-level torque commands to the joints in the arm by an operational-space impedance controller (OSC).  When training an agent using reinforcement learning, the policy is trained to output actions in the VICES action-space, directly controlling  end-effector behavior, and thus can  manage the discontinuous mode-changes made at contact using the compliance component of the space. %Additionally, since the policy is formulated in a space agnostic to the robot, it can be transferred between robots or from simulation to deployment in the physical world.

%Paragraph 2: OSC variable impedance control (VICES) is unsafe.
%\begin{itemize}
%    \item The agent loses the ability to reason about the rest of the arm, which leads to collisions + hyper-extending joints, which are both undesirable.
%    \item We can penalize this behavior in the environment (this is what Martin-Martin did), but doing so is not effective (our plots will show this). Arms break in real-life, and don't have a chance to relearn again.
%\end{itemize}

While the VICES modality serves as a useful layer of abstraction for learning and transfer, 
%an agent with an arm with more than six degrees of freedom loses the ability to reason about the configuration of the rest of its arm. 
the agent loses the ability to reason about the configuration of the rest of its arm in relation to the environment. This can lead to routine hyperextension and collisions during and after training, which are undesirable and potentially dangerous. The problem of \textit{safe} variable impedance control has therefore not yet been solved robustly \cite{2020AbuDakkaImpedanceReview}.

%Paragraph 3: Riemannian motion policies gives us a way to address the lack of safety.
%\begin{itemize}
%    \item Ratliff et al. and Cheng et al. introduce Riemannian Motion policies, a generalizes operation-space control by allowing us to specify behavior in different `operation' spaces, and then fusing them together.
%    \item RMPs have a mechanism (the Riemannian metric tensor) to encode the priority of each behavior based on the current state (position, velocity) of the arm. 
%    \item Learning the way RMPs composed has been done - RMPFusion, but learning the primary attraction behavior has not.
%\end{itemize}
In this paper, we study the problem of safely learning contact-rich manipulation behavior. 
We present RMP-VICES, a principled way to incorporate well-defined collision and joint-limit avoidance behavior into the robot's control system prior to the start of learning.
We leverage Riemannian motion policies (RMPs), which allow a user to specify different behaviors in more convenient `task' manifolds and then synthesize them  \cite{2018RatliffRMP, 2018RChengRMPFlow}. 
Most importantly, RMPs have a mechanism (the Riemannian metric of these task manifolds) for designating the priority of these behaviors, which can vary based on the current position and velocity of the arm.
We reformulate the VICES impedance controller as an attraction-type Riemannian motion policy, and synthesize it with pre-defined obstacle avoidance %defined in the distance spaces between points sampled on the meshes of the arm and the surrounding environment, 
and a joint limit behavior \cite{2018RChengRMPFlow}. 
%
%
%Paragraph 4:
%\begin{itemize}
%    \item We reformulate Martin-Martin's VICES as an attraction-type policy in the RMPTree.
%    \item We sample collision control points on the other links of the arm, and then synthesize repulsive behaviors from between those points and obstacles in the environment. These enter the RMP-tree as well.
%    \item First time someone has tried to learn a manipulation behavior scaffolded in an RMP (dispute RMPFusion)
%    \item review paper states that avoiding collisions so far relies on reachability sets, but very this is very expensive \cite{10.3389/frobt.2020.590681} \cite{safe_set_prob}
%    \item Rather than constructing the space we can enter, just repulse away from the stuff we cannot - much easier to compute with perceptuion.
%    \item The RMP-tree serves as a generalization of VICES - without the collision avoidance and joint limit policies, we obtain the original formulation of the VICES action-space controller.
%    \item Tested and verified in the simulation domains defined in VICES
%\end{itemize}
%

We verify the efficacy of our approach by demonstrating that RMP-VICES is comparably performant to state-of-the-art learning-based manipulation algorithms while largely preventing unsafe behavior. We do so by training a 7DOF Kinova Jaco 2 on three different simulated domains and analyzing the frequency and severity of collision and joint limit events that occur over the course of learning. 
Afterwards, we run the trained agents in the same domains with randomly-placed obstacles to evaluate RMP-VICES's ability to adapt to obstacles not present in training.  RMP-VICES reduces the number of collision by up to $\sim 50\%$ and joint limit events by up to $\sim 90\%$ in all of these domains with only a moderate impact on task performance. 

%===============================================================================

\begin{figure}[t]
   \begin{subfigure}{0.15\textwidth}
        \includegraphics[width=\linewidth]{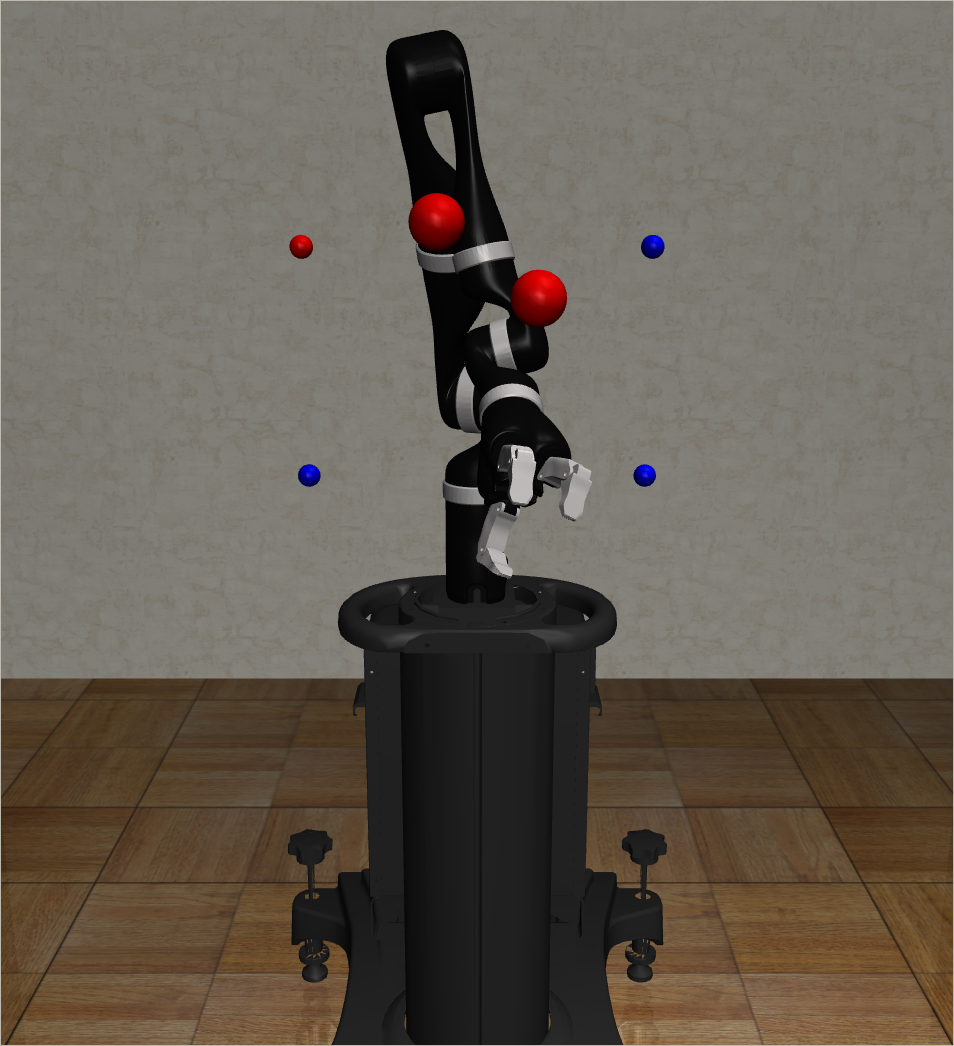}
        \caption{Traj-follow. env.}
        \label{fig:traj_col_env}
    \end{subfigure}\hfil % <-- added
    \begin{subfigure}{0.15\textwidth}
        \includegraphics[width=\linewidth]{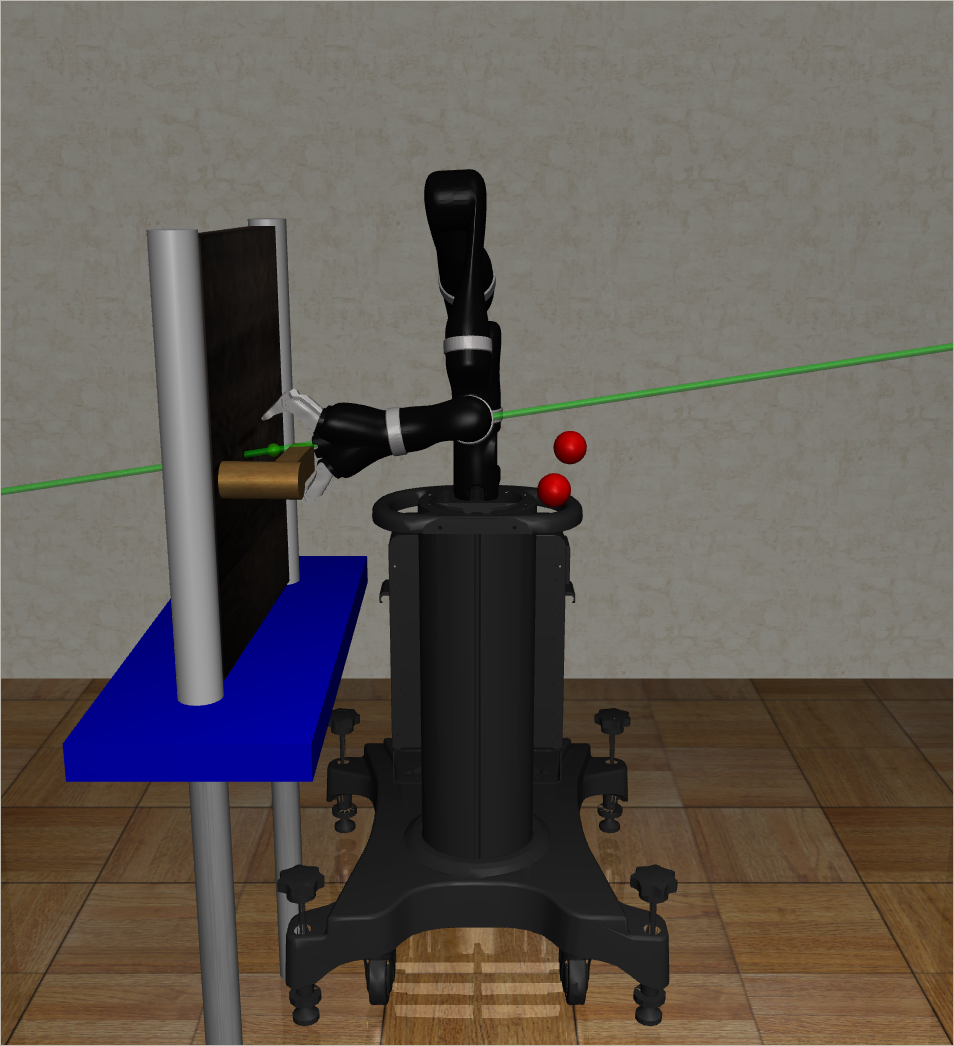}
        \caption{Door-open. env.}
        \label{fig:door_col_env}
    \end{subfigure}\hfil % <-- added
    \begin{subfigure}{0.15\textwidth}
        \includegraphics[width=\linewidth]{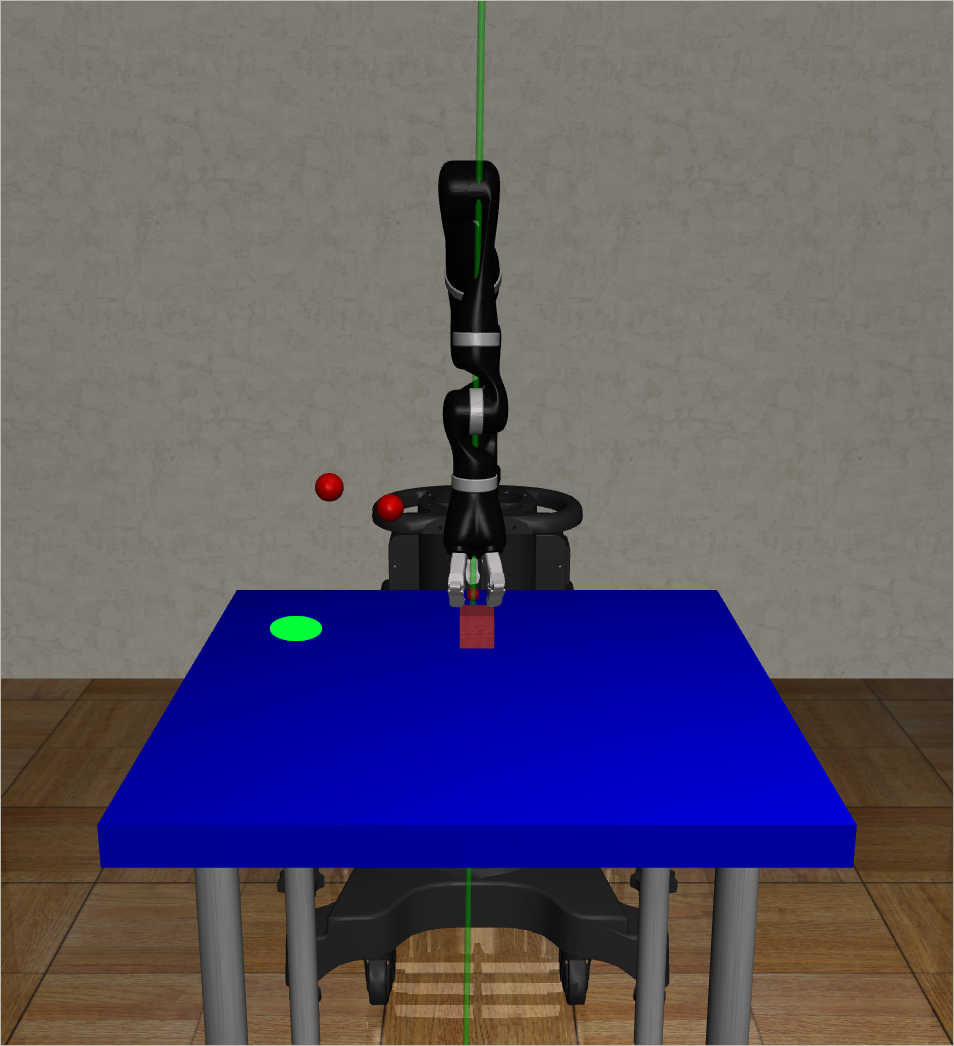}
        \caption{Block-push. env.}
        \label{fig:push_col_env}
    \end{subfigure}
    \caption{Instances of task environments with randomly-generated obstacles (red spheres).}
     \label{fig:environment_renders}
\end{figure}
\section{Background}

%  I would strongly consider restructuring this section. First, do RL! That's the problem setting. Then talk about how this is typically implemented on a robot using a policy class that can represent an expressive family of robot motions, while supporting important properties like stability. One such class is the RMP ... then go to that section! THEN at the end of the RMP section, talk about how the space we really want to do control in is OSC

\subsection{Reinforcement Learning}

% the task spaces and maps we care about - this will feed into the next section (may have to rework the ordering)
A Markov Decision process $\mathcal{M}$ is given by a tuple $(\mathcal{S, A, R, T, \gamma})$. An agent receives scalar reward $r_t$ for taking action $a \in \mathcal{A}$ in state $s \in \mathcal{S}$ according to $\mathcal{R}(s,a)$. The state evolves according to the transition function $\mathcal{T}(s,a)$; in our setting $\mathcal{S}$ and $\mathcal{A}$ are continuous spaces. The goal of reinforcement learning is to find a policy $\pi$ which maximizes the expected discounted return
\begin{equation}
    \max_\theta \mathop{\large{\mathbb{E}}}_{\mathcal{M},\pi_\theta} \left[ \sum_{t} \gamma^t r_t \right] \label{eq: rl},
\end{equation}
where $\gamma$ is a discount factor. Policy search methods are a family of algorithms that search directly over a space of parameterized policies. 
% In practice, the class of parameterized policies is typically chosen such that it can represent an expressive family of robot motions while supporting important properties such as stability. 

% since we need to discuss action spaces to fully explain vices, we placed VICES last here.

\subsection{Variable Impedance Control in End-Effector Space}

 VICES enables agents to use end-effector space ($\mathsf{SE(3)}$) and compliance of the end-effector as the action space of the agent. More formally, VICES defines an action $a = (\Delta x, k^p_{pos}, k^p_{ori}) \in \mathcal A = \mathsf{SE(3)} \times \mathbb R^3 \times \mathbb R^3$, where $k^p_{pos}$ and $k^p_{ori}$ are vector representations of stiffness matrices in an impedance controller.

In a single timestep, $\Delta x$ is composed with the current end-effector position to produce desired setpoints $p_{des}$ and $R_{des}$. The corresponding joint torques $\tau$ are then computed using Khatib's formulation of operation space control \cite{Khatib95inertialproperties, 1987KhatibOSC}:

\begin{equation}
\begin{aligned}
    \label{eqn: vices_osc}
    \tau = & J^T_{pos} [\Lambda^{pos}[k^p_{pos} (p_{des} - p) - k_v^{pos}v]] + \\
       & J^T_{ori} [\Lambda^{ori}[e_r(R_{des}, R) - k_v^{ori} \omega]],
\end{aligned}
\end{equation}

\noindent where $J_{pos}$ and $J_{ori}$ are the Jacobians associated with the FK map to end-effector position $p$ and orientation $R$, $\Lambda^{pos}$ and $\Lambda^{ori}$ the corresponding end-effector inertia matrices, and $v$ and $\omega$ the velocity and angular velocity of the end-effector. We compute damping matrices $k_v^{pos}$ and $k_v^{ori}$ so that the system is critically damped relative to given $k^p_{pos}$ and $k^p_{ori}$ as chosen by the agent. $e_r$ is an error on $\mathsf{SO(3)}$ that can be used for impedance controllers \cite{1980LuhSO3Error}; see section \ref{sec: imp_cntrl_as_rmp}.

The greatest strength and weakness of VICES is that the control signal (actions) $a$ and states are restricted to the end-effector space, $\mathsf{SE(3)}$. While this means that the agent will learn policies that can be transferred from one robot to another using $\mathsf{SE(3)}$ as a layer of abstraction, the policy is not able to reason about its joint states (and by proxy, the pose of the robot's links) that lead to its end-effector configuration to avoid collision.

\subsection{Riemannian Motion Policies}
The Riemannian motion policies framework (RMPs) is a mathematical formalism that decomposes complex robot motion behavior into individual behaviors  specified in more interpretable task spaces \cite{2018RatliffRMP}. RMPs enable the controllers to be expressed in their appropriate task spaces and then manage the transforms and flow of control between those spaces to the robot's configuration space, which is where control must ultimately occur. Additionally, they allow us to blend multiple such controllers to, for example, reach to a point (an attractive controller between the robot's end-effector and a target location) while avoiding obstacles (a repulsive or dissipative controller between the robot's arm and obstacles in its 3D environment).

RMPs are organized in a tree-like structure---the RMP-Tree---where the root represents the robot's configuration manifold $\mathcal Q$, and every node represents a task manifold (e.g. $\mathsf{SE(3)}$). The edges of the tree represent maps from the positions in the parent task space to the child task space (for an example, see Fig. \ref{fig:rmp_tree}). Here, we denote a parent space as $\mathcal M$ and the $i$th child space as $\mathcal N_i$. We will write the map from $\mathcal M$ to $\mathcal N_i$ as $\phi_i$.

Individual component behaviors to be synthesized are represented by maps from position and velocities to forces at the leaf nodes. If $\mathcal N$ is a leaf task-space, we notate the behavior map as $f(x, \dot x)$. The priority of each of these behaviors is specified by a positive-semidefinite tensor that resembles the Riemannian metric of these task manifolds,  which can vary based upon the position and velocity in that task space. We denote this metric as $M(x, \dot x) \in \mathbb R^{n \times n}$, where $n = \dim \mathcal N$, and $x \in \mathcal N$.

RMP-trees are evaluated in two phases: pushforward and pullback. In pushforward, the current state of the arm $(q, \dot q)$ is propagated through the tree to compute the state in each child manifold. In pullback, forces $f_i$ are evaluated at each child manifold and then synthesized back to compute the joint accelerations $\ddot q$ at that timestep.

\subsubsection{Pushforward}
Let $\mathcal M$ be a parent space, and let $\mathcal N_i$ be one of its child spaces, and let $x, \dot x$ be a computed position and velocity in $\mathcal M$. In pushforward, the corresponding position $y$ and velocity $\dot y$ in $\mathcal N$ are computed as $y = \phi_i(x)$ and $\dot y = J_{\phi_i} \dot x$ respectively, where $J_{\phi_i}$ is the Jacobian of task-mapping $\phi_i$. 

%Here are two key examples:

%\begin{itemize}
%    \item $\mathcal M = \mathcal Q$, i.e. joint space, and $\mathcal N = \mathsf{SE(3)}$, i.e. 6DOF free space. $\phi$ is the forward-kinematics map from joint positions and velocities to a point on the arm in free space.
%    \item $\mathcal M = \mathbb R^3$, representing the position of a point on the arm, and $\mathcal N = \mathbb R$, representing the distance space between the point and an obstacle. $\phi$ is the distance function from that point to the obstacle.
%\end{itemize}

\subsubsection{Pullback}
After pushforward is complete, we have position and velocity in each of the leaf task manifolds. We then must evaluate each policy $f_i$ and propagate the force signal back to the configuration manifold. Given parent manifold $\mathcal M$ and child leaf manifold $N_i$, with position $y_i$ and velocity $\dot y_i$ in $\mathcal N_i$ we first evaluate $\ddot y_i = f_i(y_i, \dot y_i)$ to find the corresponding leaf's acceleration, and $M_i(y_i, \dot y_i)$ to find the Riemannian-metric priority tensor. We then compute the corresponding acceleration $\ddot x$ and metric $M$ in $\mathcal M$ as follows:
\begin{align*}
    \ddot x &= \sum^n_{i=1} J_{\phi_i}^T(\ddot y_i - M_i(y_i, \dot y_i) \dot J_{\phi_i} \dot x), \\
    M &= \sum^n_{i=1} J_{\phi_i}^T M_i(y_i, \dot y_i) J_{\phi_i}.
\end{align*}

\noindent We then recursively pullback from the leaf nodes to the root node and find the joint accelerations by computing $\ddot q = M^\dagger f$, where $M^\dagger$ is the Moore-Penrose pseudo-inverse of $M$. This process solves the least-squares problem that trades off policy outputs with respect to each metric $M_i$. 

%It is important to note that while RMPs may reference the null-space implicitly by the placement of collision-avoidance policies at intermediate links between the end-effector and the base of the robot, RMPs themselves are not explicitly performing null-space optimization or control.

\subsection{Related Work on RMPs}
    %Paragraph 4: Related RMP work in manipulation. 
    
    %RMPs have been as policy classes to solve other manipulation problems \cite{DexPilot, 2019arXivRLDS, 2020arXivGUAPO}. DexPilot \cite{DexPilot} uses RMPs to enable the arm to mimic the pose of an arm of a human teleoperator to perform manipulation tasks. GUAPO does use RMPs to achieve contact-rich manipulation tasks, but only leverages the RMP for kinematic behavior in approaching the desired manipulation target. Our approach formulates the impedance manipulation behavior in the RMP itself, removing the need to identify mode changes and/or switching to a grasping controller. 
    
RMPs build on many earlier methods that decompose behavior into a number of different task spaces. Nakamura et al. \cite{1987NakamuraNullspacePriority}, Sentis et al. \cite{2005SentisSynthesis}, and Coelho and Grupen \cite{1997GrupenGrasp} all propose similar frameworks that decompose the robot task space in recursively-defined nullspaces. RMPs are formulated in a way that subsume these methods. 
% Furthermore, RMPs can have more than one behavior per space with appropriate Riemannian metric design, providing a richer set of tools to describe robot behavior.

RMPs have been used primarily to guide robot arms through free space in a variety of reaching and manipulation-based tasks \cite{li2021rmp2, 2019MukadamRMPFusion, 2020LeeGUAPO, 2020HandaDexPilot, 2018KapplerRealtime}. However, none of them perform contact-rich manipulation by training an agent whose actions are translated into attraction-type behavior in the RMP-tree itself. While not addressing contact-manipulation problems, Li et al.  \cite{li2021rmp2} introduces $\text{RMP}^2$, a framework that allows an agent to learn with RMPs to accomplish 2D reaching tasks safely using state-of-the-art deep RL methods. 

\subsection{Related Work on Safe End-Effector Variable Impedance Control}

    %Paragraph 1: Alternative approach: safe sets
    
    %\begin{itemize}
    %    \item Wabersich et al. \cite{safe_set_prob} and Wei et al. \cite{energy_funcs_and_cbf} propose a method to compute safe sets where an arm can exist without collisions.
    %    \item After computing safe sets, we can ensure arm stay in these sets with control barrier functions + other potential gradient methods (Wei et al. \cite{energy_funcs_and_cbf}, Cheng et al. \cite{cbf_funcs})
    %    \item Safe sets are expensive to compute. Rather than explicitly constructing this set, we designate 'unsafe' obstacles and construct an RMP-tree to avoid them, which is a much easier problem and boils down to perception.
    %\end{itemize} 

    An alternative approach to RMPs is to compute safety sets, and ensure that the policy stays within those bounds. Wabersich and Zeilinger \cite{2018WabersichSafeSetProb} propose a general formulation of a learning problem where these bounds can be computed. Once the safety set is known, a method such as control barrier functions \cite{2019WeiCBFs, 2018ChengCBFs} can be used to ensure that the policy stays within that set (termed \textit{forward set-invariance}). However, finding these safety sets is computationally expensive, while RMPs can be constructed from known obstacle pose data at runtime without any prior computing other than setting up the tree. While these methods are compelling due to their formal safety guarantees, we have yet to see them applied for safety of an impedance-control scheme in learning to solve contact-rich manipulation tasks.

    Many have discussed formulations of impedance control to ensure Lyapunov stability. Khader et al. \cite{2021KhaderLyapunovRL} constrain the stiffness parameters of the action space to ensure that their controller is Lyapunov-stable, and thus resilient to perturbations and unexpected behavior. Others prove Lyapunov-stability for more general formulations of learning problems  \cite{2018ChowLyapunov,2017BerkenkampLyapunovRL}. Lyapunov-stability guarantees a convergence property of the system showing that the system state must be contained in smaller regions of the space as dynamics evolve in accordance to the Lyapunov function. However, since Lyapunov-stability discusses \textit{convergence} and not \textit{safety-set forward-invariance}, these methods are not evaluated for safety using the same metrics used here.

    %Paragraph 2: 
    
    %\begin{itemize}
    %    \item Many have tried to formulate impedance control in a way that is Lyapunov stable by constraining the stiffness and damping matrices (Khader et al. \cite{Khader_lyapunov_rl})
    %    \item Other more general formulation ensure Lyapunov stability during learning 
    %(Chow et al. \cite{chow_lyapunov},  Berkenkamp et al. \cite{Berkenkamp_lyapunov_rl}). 
    %    \item These methods are not incompatible with our current design, and can actually be used together for a controller than maintains a stability guarantee as well.
    %\end{itemize}
%===============================================================================

\section{RMP-VICES}
% --
% VICES needs an OSC controller to translate high-level SE(3) commands from policy to low-level torques. We are replacing that controller with something that is more `geometrically aware. (RMP)' 
%
%
%

A naive operational-space controller (e.g. as in VICES \cite{1987KhatibOSC}) fails to reason about collision with the environment and violation of the arm's joint limits. We propose RMP-VICES, an operational-space controller which translates $\mathsf{SE(3)}$ impedance commands from a learned policy to low level torques that accounts for the geometry of the configuration space and surrounding environment. 
We formalize variable-impedance control as an attractor-type RMP, and fuse it with repulsion-type RMPs based at points sampled from the robot's arm. 
The resulting controller blends the desired policy control with collision avoidance behavior in training and in deployment.

In this paper, we assume perfect knowledge of the robot arm dynamics and the pose and geometry of all obstacles in the environment, but neither object nor contact dynamics. The goal of RMP-VICES is to solve manipulation tasks with comparable performance to VICES but with fewer obstacle collisions and joint-limit extensions over training and deployment.

\subsection{Tree Structure}

The forward-kinematics (FK) map to the last link of the arm is used to compute the pose of the end-effector in $\mathsf{SE(3)}$. We then perform a selection mapping to decompose $\mathsf{SE(3)}$ into $\mathbb R^3$ and $\mathsf{SO(3)}$, and define the variable impedance behavior in those spaces. We perform an identity map from the robot's configuration space to itself, where we define the joint-limiting behavior as shown in Cheng et al. \cite{2018RChengRMPFlow} and was tuned via experimental validation.

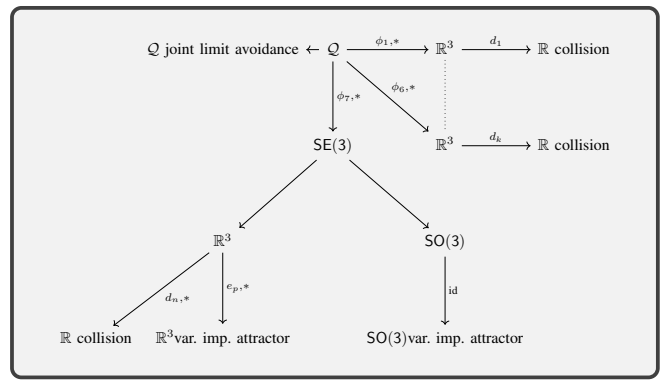
\begin{figure}[ht]
\begin{tcolorbox}
\adjustbox{scale=0.60,center}{
% https://tikzcd.yichuanshen.de/#N4Igdg9gJgpgziAXAbVABwnAlgFyxMJZAJgBoAGAXVJADcBDAGwFcYkQAdDgW3pwAsAxk2ABFAL4hxpdJlz5CKMgEZqdJq3ZdeAuADNgAZQCiACgDMASknTZ2PASLLSxNQxZtEnHn34AjP2AAJXEAPXMpGRAMewUicxc3DU9vHX59IwB5C2tIu3lHFGdzJI8tHwEA4LDzAAIuHBgADxxgBgAnLHo-RhharG40GCh6MEE+vhx2+kEcCHabKJiCxWQEkpp3TS9tXwzDbKsa+o5Glrb6Tu7e-sHh0fHayenZ+cX8h1WE1U3k8rSqiFwnlonJPvEKKVtqlfICaiDluCUAAWSG-Mo7Cr+QIhE5nVq1QQQRiMLDYAjvUGxQrIVE-dQYmGVHHiPHNAlEklkhSUxFxFDkUgbBnQ3bM6ps86E4mk8lgXlg-nIZxUdGirHCRi1USSgkAKwgWDAOFqpO4uCetENIzGMBsamGAHN4ERQHp2hBuEhBSA5khnCBGN0YIwAAqKwqBmB6HAgNUpLhofhYAD6AHZSLUAFQg92e-00P2IYi2EB5r2IANF8yl8tIMi+iBIVEilIwFNoTM52seis+osANho-Bg9CgSDAzBJND8MDA48Q-foWEY7EgYDY8fKSdTQ+zud73sLTcQGdb2+TKec+57+dPx6QQ5AI7Ha4Im8DRpSUAgOEa49vCszyLAAOLcvCgK8DzvJ8iwATnAkBIIAa2gisGyLZQfS2b8UzALs0KQBJGyQABWRCGnZYAsCgSk60XB9K2UcRKHEIA
\begin{tikzcd}[column sep={0.010cm},row sep={1.50cm}]
                             & \mathcal Q \text{ joint limit avoidance}                  & \mathcal{Q} \arrow[d, "{\phi_7, *}"] \arrow[rd, "{\phi_6, *}"] \arrow[r, "{\phi_1, *}"] \arrow[l] & \mathbb{R}^3 \arrow[d, no head, dotted] \arrow[r, "d_1"] & \mathbb{R} \text{ collision} \\
                             &                                                           & \mathsf{SE(3)} \arrow[ld] \arrow[rd]                                                              & \mathbb{R}^3 \arrow[r, "d_k"]                            & \mathbb{R} \text{ collision} \\
                             & \mathbb{R}^3 \arrow[d, "{e_p, *}"] \arrow[ld, "{d_n, *}"] &                                                                                                   & \mathsf{SO(3)} \arrow[d, "\text{id}"]                    &                              \\
\mathbb{R} \text{ collision} & \mathbb{R}^3 \text{var. imp. attractor}          &                                                                                                   & \mathsf{SO(3)} \text{var. imp. attractor}     &                             
\end{tikzcd}
}
\end{tcolorbox}
\caption{The RMP-tree structure used for RMP-VICES. $\phi_1,..., \phi_7$ denote FK maps from $\mathcal Q$ to each link of the arm, where VICES sits in $\mathsf{SE(3)}$. The rest of the RMP-tree is made up to control points sampled from the link meshes, which are mapped to the shortest-distance space each obstacle ($d_1, ..., d_n$ act as signed-distance functions) where collision avoidance is defined. The selection map from $\mathsf{SE(3)}$ to $\mathbb R^3$ has been omitted.}
\label{fig:rmp_tree}
\end{figure}

For the purpose of collision avoidance, we sample control points on the links of the arm, and define distance spaces to each obstacle.
Since we conducted training with a 7DOF arm, we need seven different FK maps from the configuration space to $\mathsf{SE(3)}$ for the pose of each individual link of the arm.
From each of the sampled control points, we compute a selection map to $\mathbb R^3$ and then a distance map to $\mathbb R$ for each obstacle, and define collision-avoidance behavior there. A schematic of the RMP-tree can be seen in Fig. \ref{fig:rmp_tree}.

\subsection{Impedance Control as Attractor-type RMPs} 
\label{sec: imp_cntrl_as_rmp}
We formulate impedance controllers as RMP leaves in the $\mathbb R^3$ and $\mathsf{SO(3)}$ task spaces. Both of these controllers are formulated as attraction-type policies \cite{2018RatliffRMP}. 

%While the formulation of an impedance controller n $\mathbb R^3$ can naturally be expressed as a GDS, the same cannot be said about $\mathsf{SO(3)}$. Thus, we present a variation of the typical $\mathsf{SO(3)}$ impedance formulation to frame it as a GDS in the RMP.

Let $x \in \mathbb R^3$ be a position of the end-effector in free space and $\dot x$ be its associated velocity as computed by the pushforward operation. Let $x_g \in \mathbb R^3$ be a desired position. Then the motion behavior is formulated as a spring-damper attractor-type policy in $\mathbb R^3$:
\begin{equation}
    \label{eqn: impedance_pos}
    f(x, \dot x) = k_p (x - x_g) - k_d (\dot x),
\end{equation}
with the associated metric being the identity matrix $I$. The diagonal matrices $k_p^{pos}$ and $k_d^{pos}$ specify the stiffness of this controller and are set by the trained policy at every timestep.

The attractor in $\mathsf{SO(3)}$ is written in a very similar way to the RMP above, except we also must account for the non-Euclidean topology of $\mathsf{SO(3)}$ in the error term. Let $r \in \mathsf{SO(3)}$ and $\omega$ be the end-effector's current orientation and angular velocity, and $r_g$ be the desired orientation. We represent $r, r_g \in \mathsf{SO(3)}$  as rotation matrices, i.e. $r = [r_x, r_y, r_z]$ and $r_g=[r_{gx}, r_{gy}, r_{gz}]$.
% $$
%     r =
%     \left[
%     \begin{array}{ccc}
%         \vertbar & \vertbar & \vertbar\\ 
%         r_x     & r_y & r_z \\
%         \vertbar & \vertbar & \vertbar
%     \end{array}
%     \right]
%     \mbox{, }
%     r_g =
%     \left[
%     \begin{array}{ccc}
%         \vertbar & \vertbar & \vertbar\\ 
%         r_{gx}     & r_{gy} & r_{gz} \\
%         \vertbar & \vertbar & \vertbar
%     \end{array}.
%     \right]
% $$
We then use the same error term on $\mathsf{SO(3)}$ from VICES:
\begin{equation}
    \label{eqn:ori_error}
    e_r(r, r_g) =  \frac{1}{2}\left(r_x \times r_{xg} + r_y \times r_{yg} + r_z \times r_{zg}\right).
\end{equation}
\noindent
This error is equivalent to the sine of the angle to rotate $r$ to $r_g$ about a single axis \cite{1980LuhSO3Error}. We then define the impedance controller using this error:
\begin{equation}
    \label{eqn:ori_impedance}
    f(r, \omega) = k_p^{ori} \cdot e_r(r, r_g) - k_d^{ori} \cdot \omega,
\end{equation}
 where $k_p$, $k_d$ are the stiffness and damping coefficients respectively. As before, the metric associated with this attraction-type policy is the identity $I$.

It is important to note that removing the obstacle-avoidance and joint-limit policies from the RMP-tree reduces the controller to VICES \cite{2018RatliffRMP, 2019MartinMartinVICES}.

\subsection{Collision Avoidance RMPs}
For both types of obstacles, we use the signed-distance function to map the position of the control point sampled on the arm to the distance space between the point and the obstacle. We define the Riemannian metric $m(x, \dot x) \in \mathbb R$ of this 1-dimensional space to be the following expression:
\begin{equation}
    \label{eqn:collision_metric}
    m(x, \dot x) = 
        \frac{(\max \{\dot x, 0\})^2}{x^4}
\end{equation}
This metric was derived via experimental validation. As described in \cite{2018RatliffRMP}, the collision policy will only activate and strengthen if the control point is moving towards the obstacle. %The degrees of the numerator and denominator were validated experimentally.
%For the spherical obstacles, the collision behavior $f$ is defined to be the gradient of an artificial potential function $\Phi(x)$ to generate a repulsive behavior:
% \begin{equation}
%     \label{eqn:potential_collision}
%     f(x, \dot x) = \nabla \Phi(x) = \frac{-4\alpha}{x^9}.
% \end{equation}
% where $\alpha$ is a tuning coefficient. Both the repulsion and metric functions are inspired by the collision policies defined in \cite{2019aAnqiMultiRMP}. Like the Riemannian metric, the exponents were determined experimentally.
% discuss the presence of potentials for spherical obstacles, but not for planes, and then give a closed formulation of the potential.
For both planar and spherical collision-avoidance behaviors, we do not provide a repulsive signal but a damping term ($f(x, \dot x) = \eta \dot x$, for some damping coefficient $\eta$). As Bylard et al. in \cite{2021BylardPBDS} observe, using only the Riemannian metric and a dissipation function to reduce the movement towards the obstacle to generate collision-avoidance behavior reduces the likelihood of the arm to trap itself in local-minima.

%\begin{equation}
%    \label{eqn:jlim_null}
%    f(q, \dot q) = \eta_p (q_0 - q) - \eta_d \dot q
%\end{equation}

%where $q_0$ is initial position of the arm, and $\eta_p$ and $\eta_d$ are the proportional and derivative gains respectively. The Riemannian metric is a tensor whose entries scale up when an individual joint is (1) moving towards the joint limit and strengthens as the joint moves further towards the limit, which scale the contribution of the behavior to the overall outpout in a similar way to the metric for our collision avoidance RMPs. Equation \ref{eqn:jlim_null} also serves to resolve the nullspace of our 7DOF arm.

\subsection{Integrating Policy Actions in the RMP-VICES Controller}

% \begin{figure} %{r}{0.5\textwidth}
%     \centering
%     \includegraphics[width=0.50\linewidth]{figures/rmp-loop.png}
%     \caption{Schematic of control loop. }
%     \label{fig:loop}
% \end{figure}

In an MDP formulation of a manipulation task, our state space will always be a superset of $\mathsf{SE(3)} \times S_{obj}$, where $\mathsf{SE(3)}$ is the space of end-effector pose and $S_{obj}$ is the space of states of the manipulated object. An action will be a tuple $(\Delta x, k_p^{pos}, k_p^{ori}) \in \mathsf{SE(3)} \times \mathbb R^{3} \times \mathbb R^{3}$, where $\Delta x$ is a displacement in $\mathsf{SE(3)}$ for the next goal pose and $k_p^{pos}$ and $k_p^{ori}$ are the stiffness coefficients for the impedance controllers ($k_d$ is computed so the impedance controllers are critically-damped). The policy $\pi$ will be sampled in pushforward and used to update $k_p^{pos}, k_p^{ori}$, $k_d^{pos}, k_d^{ori}$, and $r_g$, and $x_g$ in eqs. \ref{eqn: impedance_pos} and \ref{eqn:ori_impedance} in the RMP-tree on every timestep.

After the pullback stage of the RMP, we obtain a corresponding joint acceleration $\ddot q$ given by the synthesis of the end-effector impedance control and collision avoidance behavior. To convert $\ddot q$ into joint torques, we multiply by the inertia matrix of the arm in joint space and compensate for gravity and Coriolis forces.

% \subsection{RMPFusion Pullback Weighting}
% While the Riemannian metrics defined at leaf nodes policies are supposed to act as a priority weights when policy outputs are combined in pullback, designing them for proper weighting is difficult in practice. Mukadam et al. extends RMPs by incorporating a scalar weighting as a function of state in the parent space \cite{2019arXivRMPFusion}. 

% More formally: let $\mathcal M$ be a parent space, and $\mathcal U_1, \mathcal U_2, ..., \mathcal U_n$ be its child task-spaces. Let $w: \mathcal M \to \mathbb [0, \infty)^n$ be a weighting function. Then pullback is modified as follows: 

% \begin{flalign*}
%     f &= \sum^n_{i=1} w_iJ_{\phi_i}^T(f_i - M_i \dot x) + h_i\\
%     M &= \sum^n_{i=1} w_iJ_{\phi_i}^T M_i J_{\phi_i}\\
% \end{flalign*}
% \noindent
% where $w(x) = (w_1, w_2, ..., w_n)$ and $x$ the current pose in $\mathcal M$, and $h$ a specialized damping term to preserve Lyapunov-stability. The weighting function $w$ can be defined by hand, but more typically will be defined using a function approximator like a neural network. It is important to note that if $w(x) = (1,1,...,1)$, then RMPFusion reduces back to the original RMP definition.

% In the diagram above, we denote all possible branches that are good candidates for RMPFusion weighting with a star (*).

%===============================================================================

\section{Experiments}

  We test and verify the efficacy of RMP-VICES on the trajectory-following and door-opening domains constructed in Martin-Martin et al. \cite{2019MartinMartinVICES} and an additional block pushing domain we developed. In each simulated domain, we first train VICES and RMP-VICES in an environment with no additional obstacles to evaluate any negative impact the RMP-tree has on learning efficiency. We also record the force of all collisions of the arm with objects in the workspace to understand the severity of collisions events using VICES and RMP-VICES. Collision and joint-limit events incur a penalty in the reward function and termination of the episode (but an ablation on these two properties showed that they were not essential for effective training). Afterwards, we perform 100 rollouts with the learned policy in each domain with two randomly placed spherical obstacles and record the force of any collision events to verify the adaptability of each algorithm to obstructions not seen during training.

All simulated experiments were conducted in Robosuite \cite{2020ZhuRobosuite}. The agent was given a stochastic policy parameterized by two fully-connected layers of 64 nodes with $\tanh$ activations and optimized using PPO, as done in VICES. Our PPO implementation is based on the code written for OpenAI's SpinningUp \cite{2018AchiamSpinningUp}. In each domain, each episode was given a length of 1024 steps, with 4096 steps per epoch. The policy network was initialized with random weights (10 seeds) and was trained for 367 epochs ($~1.6 \times 10^6$ training steps).
%For the physical demonstration, we constructed a block-pushing task in Robosuite @\cite{robosuite2020} and trained our agent in simulation. We then reconstructed the environment and evaluated the efficacy of RMP-VICES's ability to generate trajectories that avoid adversarially obstacles placed in the workspace.

\subsection{Environments}

\subsubsection{Trajectory-Following}

We randomly place four via points in the workspace, and specify an order in which the robot end-effector must traverse through them. 
%The robot receives a small reward shaped by the distance between the end-effector and the next via-point. The agent also received an additional reward for passing through each via-point and a substantial reward for task-completion. 
As in VICES, the state of the agent is represented by the pose and the velocity of the end-effector, the position of each via-point, and whether each via-point has been checked. For the rollouts after training, we randomly generate two spherical obstacles with a radius of 5 cm in the bounding volume that is used to generate the via-points. We ensure that no obstacles overlap with a via-point so that the task still has a guaranteed solution.

\subsubsection{Door-Opening}

 %As before, the reward function gives a shaped reward as a function of the difference between the door's current hinge angle and the target hinge angle. The robot is also initialized with the end-effector near the door with the gripper's fingers closed in a cage around the handle. As before, the state of the agent is represented by the end-effector's position and velocity, as well as the current angle hinge angle of the door. 
The RMP-tree is initialized with two plane collision avoidance policies to avoid the panel of the door and the surface of the table. For the rollouts after training, we generate two spherical obstacles in a bounding volume placed at a distance of front of the door to ensure that it can still swing open for task completion.

%\subsection{Physical Demonstration}

%For the physical demonstration, we constructed a block-pushing task where the robot must push a weighted-block placed on a table to a designated position elsewhere on the table. The block's starting location is randomized, and the goal location is randomly sampled to be at most 25cm away from the starting position of the block. The state of the agent is represented by the pose and velocity of the end-effector, a vector from the end-effector to the cube, and a vector from the cube to the goal position.

\subsubsection{Block-Pushing}

We construct a domain where the robot must push a block across a table to a goal position. The state includes the end-effector's pose and distance to the block as well as the distance between the block and the goal position. The reward function gives a small shaped reward that depends on the distance between the end-effector and the goal, and a larger shaped reward between the cube and the goal:
\begin{equation}
\begin{aligned}
    r_{main}(d_{h2c},d_{c2g}) = & r_{h2c} * (1 - \tanh{(20 d_{h2c})}) + \\
    &r_{c2g} * (1 - \tanh{(20 d_{c2g})}),
\end{aligned}
\end{equation}
where $d_{h2c},d_{c2g}$ are the distance form the hand to the cube, and distance from the cube to the goal respectively. $r_{h2c}$ and $r_{c2g}$ are the maximum rewards made to provide incentive for the agent to bring the end-effector to the cube, and the cube to the goal, respectively. %Colliding with the environment or reaching joint limits heavily is heavily penalized and terminates the episode.

\begin{figure}[h]
    \centering
    \vspace{1em}
    \begin{tabular}{cc}
        \raisebox{1em}{\rotatebox{90}{VICES}} &
        \includegraphics[width=0.9\linewidth]{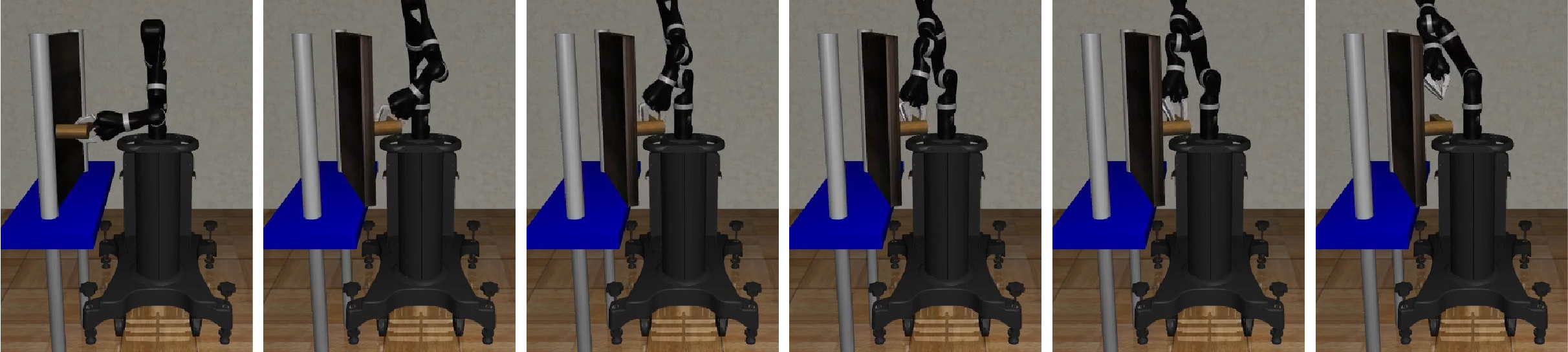} \\
        \raisebox{0em}{\rotatebox{90}{RMP-VICES}} &
        \includegraphics[width=0.9\linewidth]{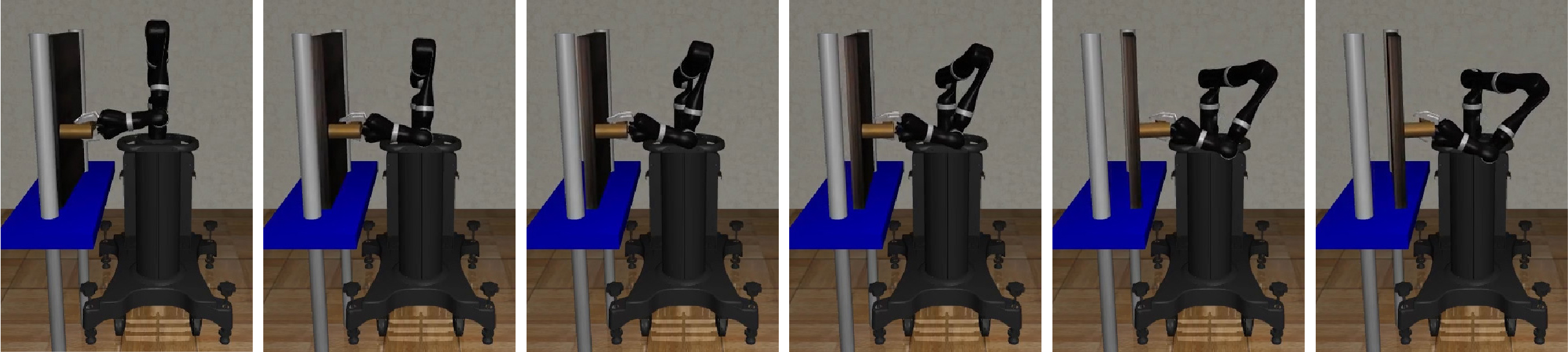} \\
    \end{tabular}
    \caption{Regularly-timed stills of a single rollout of VICES (top) and RMP-VICES (bottom) after training is complete.}
    \label{fig:door_story_board}
\end{figure}

\begin{figure}[ht]
    \vspace{-1.0em}
    \centering
    \begin{subfigure}{0.17\textwidth}
        \includegraphics[width=\linewidth]{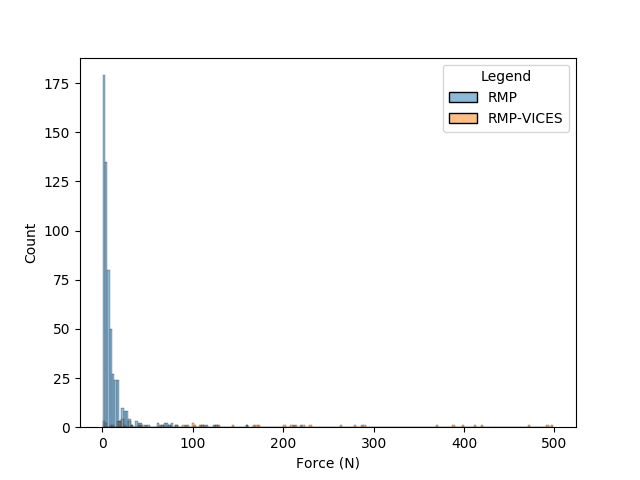}
        \caption{Door-open.}
        \label{fig:door_train_hist}
    \end{subfigure}\hfil % <-- added
    \begin{subfigure}{0.17\textwidth}
        \includegraphics[width=\linewidth]{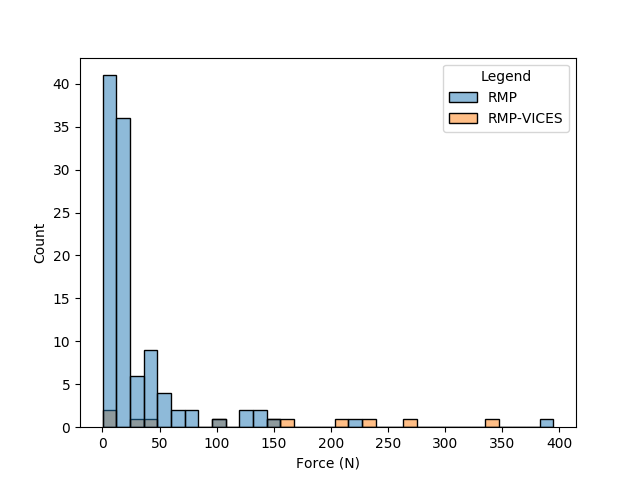}
        \caption{Block-push.}
        \label{fig:push_train_hist}
    \end{subfigure}\hfil % <-- added
    \begin{subfigure}{0.15\textwidth}
        \includegraphics[width=\linewidth]{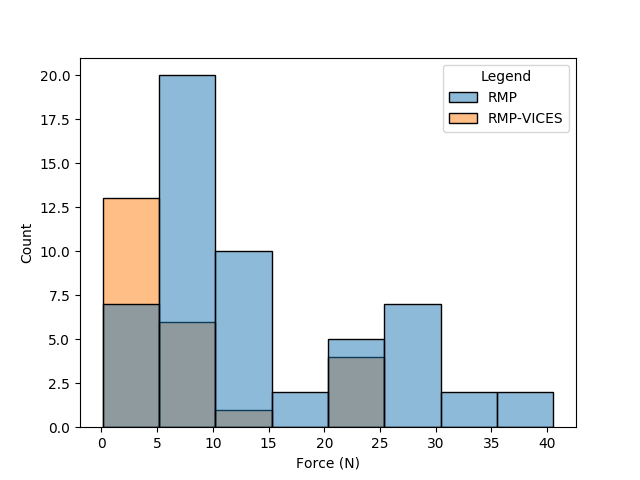}
        \caption{Traj-follow.}
        \label{fig:traj_zero_hist}
    \end{subfigure}\hfil % <-- added
    \begin{subfigure}{0.15\textwidth}
        \includegraphics[width=\linewidth]{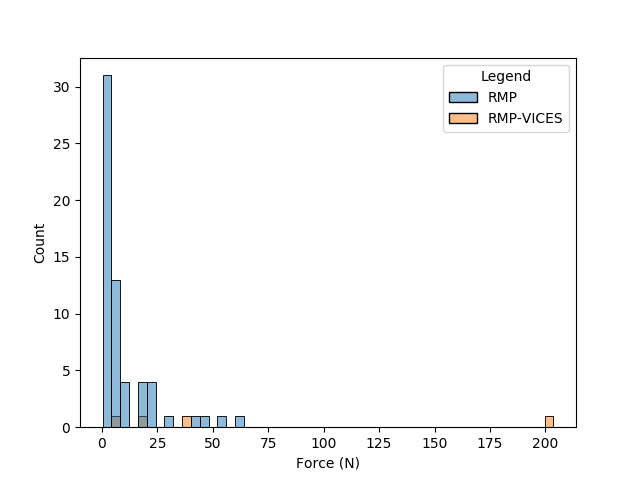}
        \caption{Door-open.}
        \label{fig:door_zero_hist}
    \end{subfigure}\hfil % <-- added
    \begin{subfigure}{0.15\textwidth}
        \includegraphics[width=\linewidth]{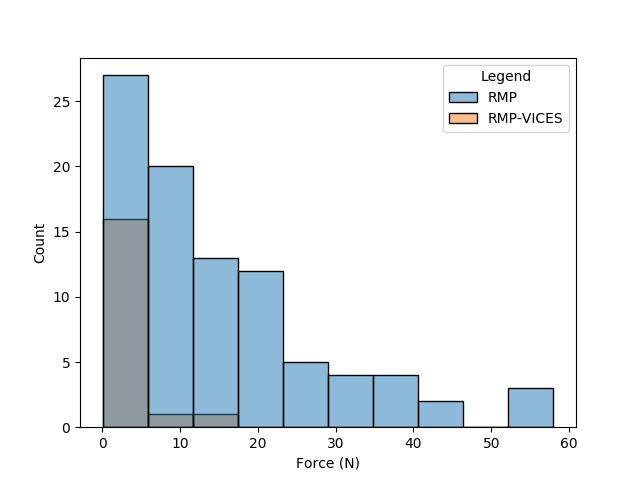}
        \caption{Block-push.}
        \label{fig:push_zero_hist}
    \end{subfigure} 
    \caption{Severity of collision events in training (a,b), and in environments with randomly generated obstacles (c-e).}
    \label{fig:training_collision_hists}
\end{figure}

\section{Results}

\begin{figure*}[ht]
    \centering
   \begin{subfigure}{0.25\textwidth}
        \includegraphics[width=\linewidth]{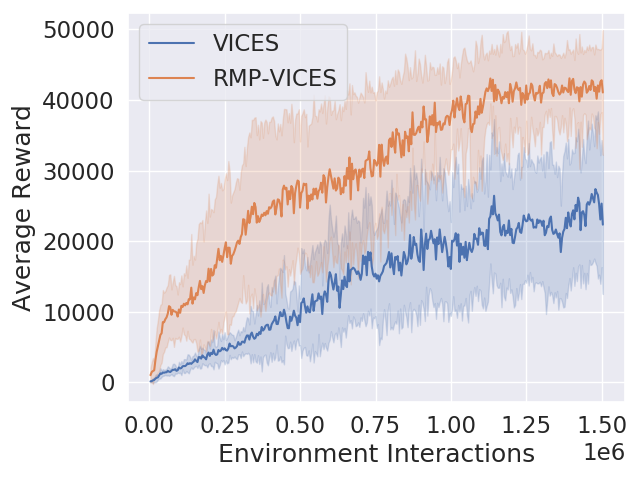}
        \caption{Traj-follow., performance}
        \label{fig:traj_perf_train}
    \end{subfigure}\hfil % <-- added
    \begin{subfigure}{0.25\textwidth}
        \includegraphics[width=\linewidth]{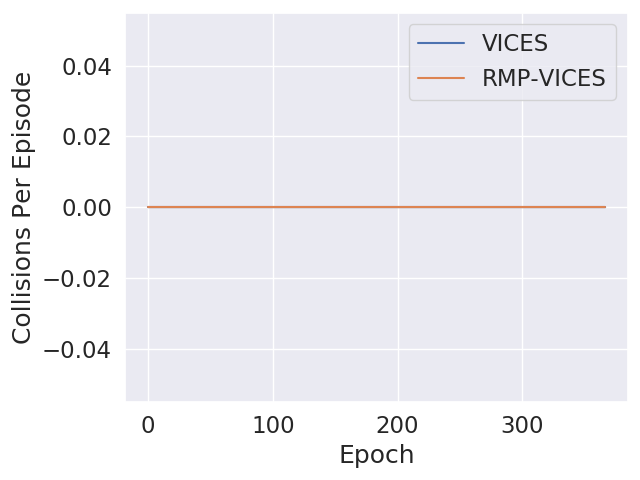}
        \caption{Trajectory-follow., collisions}
        \label{fig:traj_col_train}
    \end{subfigure}\hfil % <-- added
    \begin{subfigure}{0.25\textwidth}
        \includegraphics[width=\linewidth]{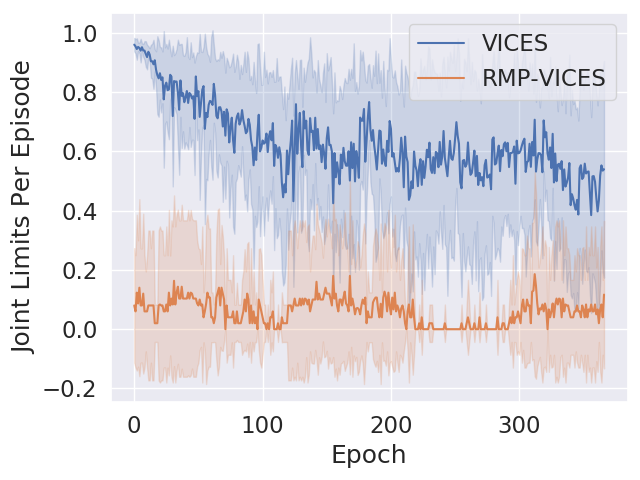}
        \caption{Trajectory-follow., joint limits}
        \label{fig:traj_jlim_train}
    \end{subfigure}
    \medskip
    \begin{subfigure}{0.25\textwidth}
        \includegraphics[width=\linewidth]{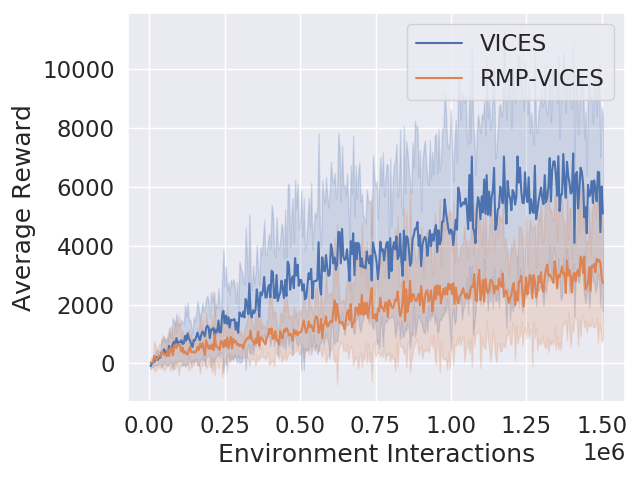}
        \caption{Door-opening, performance}
        \label{fig:door_perf_train}
    \end{subfigure}\hfil % <-- added
    \begin{subfigure}{0.25\textwidth}
        \includegraphics[width=\linewidth]{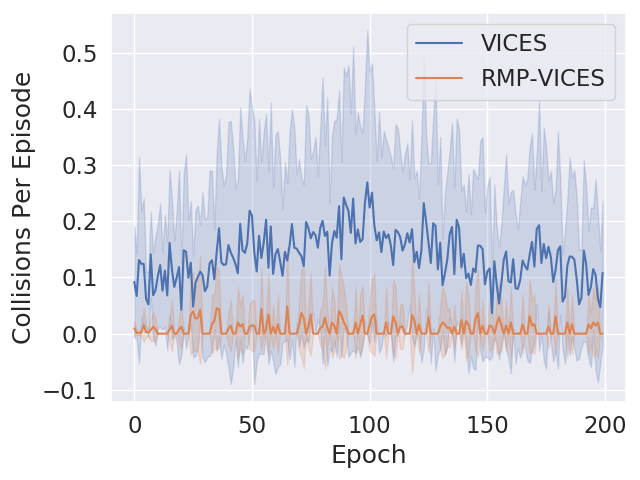}
        \caption{Door-opening, collisions}
        \label{fig:door_col_train}
    \end{subfigure}\hfil % <-- added
    \begin{subfigure}{0.25\textwidth}
        \includegraphics[width=\linewidth]{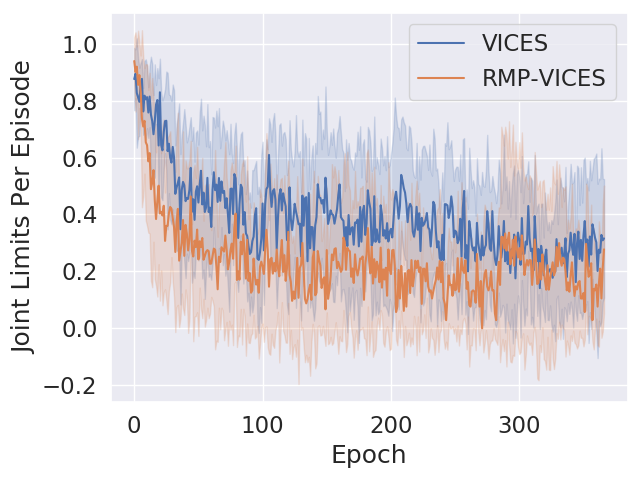}
        \caption{Door-opening, joint limits}
        \label{fig:door_jlim_train}
    \end{subfigure} 
    \medskip
    \begin{subfigure}{0.25\textwidth}
        \includegraphics[width=\linewidth]{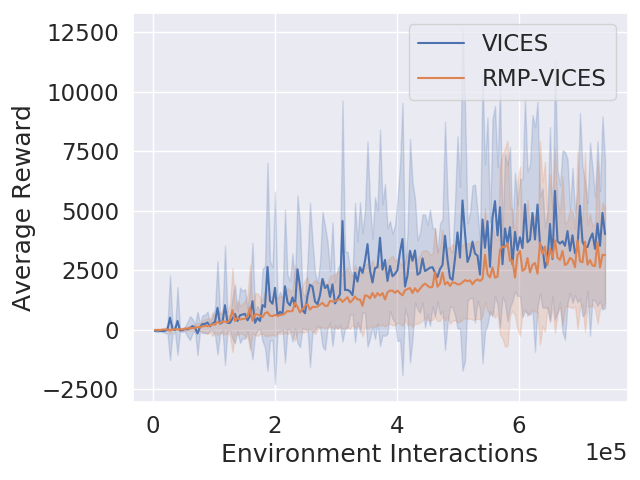}
        \caption{Block-pushing, performance}
        \label{fig:push_perf_train}
    \end{subfigure}\hfil % <-- added
    \begin{subfigure}{0.25\textwidth}
        \includegraphics[width=\linewidth]{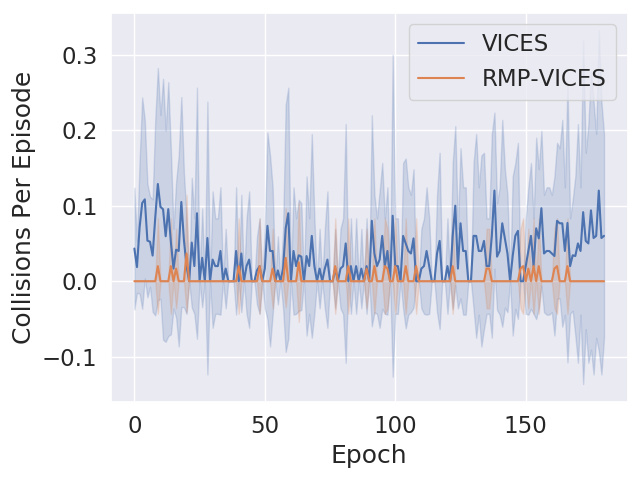}
        \caption{Block-pushing, collisions}
        \label{fig:push_col_train}
    \end{subfigure}\hfil % <-- added
    \begin{subfigure}{0.25\textwidth}
        \includegraphics[width=\linewidth]{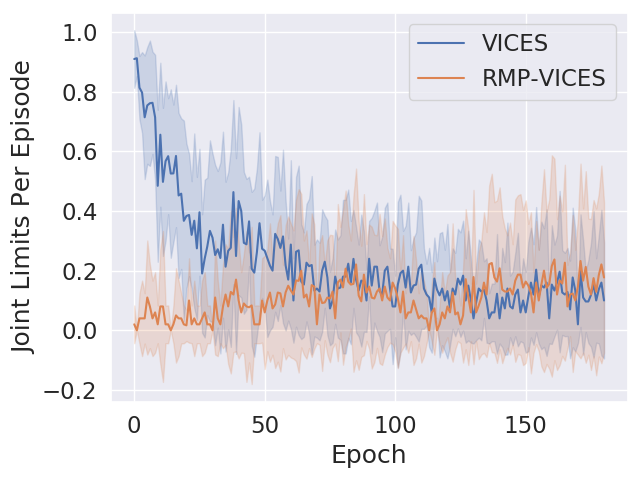}
        \caption{Block-pushing, joint limits}
        \label{fig:push_jlim_train}
    \end{subfigure} 
    \caption{Average reward, collision, and joint limit performance of VICES and RMP-VICES over the course of training on the trajectory-following (\ref{fig:traj_perf_train},\ref{fig:traj_col_train},\ref{fig:traj_jlim_train}), door-opening (\ref{fig:door_perf_train},\ref{fig:door_col_train},\ref{fig:door_jlim_train}), and block-pushing domains (\ref{fig:door_perf_train},\ref{fig:door_col_train},\ref{fig:door_jlim_train}). When integrating the average curves for joint limits and collision avoidance behaviors, we see that RMP-VICES outperforms VICES in every domain. Percent decrease in collision and joint limit rates, respectively: trajectory-following: N/A\%, 90.5\%; door-opening: 39.8\%, 90.7\%; block-pushing: 54.6\%, 93.2\%.}
    \label{fig:training_data}
\end{figure*}

\subsection{Safety During Training}
%Paragraph 1: impact of RMP on learning
%\begin{itemize}
%    \item Trajectory following: THE rmp was more sample efficient + found a better policy - avoiding joint limits injected bias that enabled the policy to learn better behavior. No collisions to be had during training. Joint limit behavior is at the level that the normal VICES control had to learn over time. 
    
%    \item Door-opening: RMP was slower learner. Likely because the arm trajectory was super close to the door, and was blocked by the RMP. Collisions are much better on RMPs - RMP took a safe root to learn a very similar policy without breaking into the door without restraint, since we have effectively limited a part of the search space. The arm gets wedParagraph 1: impact of RMP on learningged in between the door and the other obstacles, and thus gets trapped - hence collisions.
    
%    \item Block-pushing. Performance is similar. RMPs start much better. But drawback, to learn here, we see that learning picks up after the policy starts fighting the rest of the tree. Prehaps having a Lyapunov-stable controller may aid in ensuring that the policy doesn't explore erratic behavior that defeats the purpose of the RMP-tree. Can see worse collisions since the policy made more severe decisions in fighting the RMP. The RMP only gives soft limit.
%\end{itemize}

%\subsection{Influence of RMP-VICES When Learning}

In the trajectory-following task, RMP-VICES has far fewer joint-limit events than VICES (Fig. \ref{fig:traj_jlim_train})  throughout training. Furthermore,  RMP-VICES learns more sample-efficiently than just VICES alone (Fig. \ref{fig:traj_perf_train}). Since violating a joint limit causes episode termination, RMP-VICES has access to more viable data earlier in training, which accelerates the agent's ability to learn the task.

In the door-opening task, we see that while RMP-VICES  avoids collision better than VICES (figs. \ref{fig:door_col_train}, \ref{fig:door_train_hist}), the agent trained with RMP-VICES learns  with worse overall performance (Fig. \ref{fig:door_perf_train}). However, RMP-VICES is significantly safer in guiding the robot away from collision avoidance behavior. After reviewing several rollouts of each policy (Fig. \ref{fig:door_story_board}), we see that the policy learns a behavior that keeps the robot's pose largely the same with the elbow pointing straight up, which the RMP-VICES joint-limit and collision-avoidance policy represses. As a result, the RMP-VICES controller first pitches the elbow downward, and then opens the door. Since we give a dense reward that increases with door angle, the RMP-VICES agent obtained a lower average reward in VICES. However, the RMP-VICES policy is a safer behavior than the one learned by VICES alone, since the VICES behavior is prone to collision with the door (Fig. \ref{fig:door_story_board}). This problem can be addressed with better tuning of the joint-limit policy or by learning the Riemannian metric \cite{li2021rmp2}.

%Since we now are running collision avoidance, which repels the arm from away from the door and table in the environment, we have shrunk the agent's exploration space. As such, if there are efficient solutions that exist near or outside the boundary of the exploration space, an agent trained with RMP-VICES will not be able to find those solutions since it cannot reach them, as shown by few collision event in Fig. \ref{fig:door_col_train}, \ref{fig:door_train_hist}.

In the block-pushing task, we see that RMP-VICES starts with better joint-limiting performance at the beginning of training figs. \ref{fig:push_col_train} \ref{fig:push_jlim_train}. While VICES performs slightly better than RMP-VICES in efficacy of learning, RMP-VICES is much more performant in collision avoidance throughout training.

%\subsection{Efficacy of RMP-VICES in Environments with Unseen Obstacles}

%\begin{itemize}
%    \item In most instances, we can see that RMP-VICES generates fewer collisions of comparable to moderate severity. But there are two instances where it did not: when the policy began fighting back the RMP to solve the task in training, and when the policy gets wedged in between the door and the obstacles on rollouts. The two stray collisions on FreeSpaceTraj - they are 
%\end{itemize}

\subsection{Deployment in Environments with Generated Obstacles}

For each experiment, we choose the best performing seeds from the VICES and RMP-VICES agent from each domain, and played through 100 episodes with randomly-generated spherical obstacles in the environment. The size and location of the bounding volumes used for each domain were chosen as areas where the arm would frequently pass through to the solve the task. Table \ref{tab:zero_stats} gives an aggregation of the performance of VICES and RMP-VICES from this set of experiments. 
See Fig. \ref{fig:environment_renders} in the for visualizations of all three environments for a single rollout instance.

Relative to performance in training, we see a drop in performance in all tasks. This makes intuitive sense because neither the VICES nor RMP-VICES policies experienced these obstacles before training. RMP-VICES, in all tasks, collides less frequently than VICES. While we do see a slight drop in performance in average reward, it is important to note that the success of the agent must be measured by its efficacy its ability to remain safe while completing the task.

\begin{table}[ht]
\centering
\begin{tabular}{@{}lllll@{}}
\toprule
                          & Task                 & Avg. Reward                  & Coll. & Jnt Lim. \\ \midrule
\multirow{3}{*}{VICES}     & Traj-follow. & $7864 \pm 1.167 \times 10^4$   & 56                               & 44                                  \\
                          & Door-open.         & $1703 \pm 1.948 \times 10^3$  & 61                             & 36                                 \\
                          & Block-push.        & $1755 \pm 2.256\times 10^3$  & 93                               & 0                                   \\
\cmidrule{2-5}
\multirow{3}{*}{RMP-VICES} & Traj-follow. & $6847 \pm 3.723\times 10^3$   & 25                             & 22                                 \\ % spin on this one a bit: 2300 \pm 2.5 \times 10^3, 60, 11
                          & Door-open.         & $1051 \pm 6.96\times 10^2$ & 5                               & 11                                  \\
                          & Block-push.        & $1740. \pm 8.54 \times 10^2$ & 19                               & 0                                  \\ 
                          \bottomrule
\end{tabular}
\caption{Results from rollouts with randomly-generated obstacles. As collision or passing a joint limit triggers the episode to end, the maximum number of collision and joint events for a single domain for a single agent is 100.}
\label{tab:zero_stats}
\end{table}

It is clear that the agent trained on RMP-VICES exploits the geometric prior given by the collision-avoidance RMPs for more safe behavior. An agent trained with VICES alone has no obvious way to incorporate this information, and the arm moves through the space blindly without adjusting its trajectory to avoid the new obstacles in its path.

Furthermore, we see that RMP-VICES reduces the impact of collision between the arm and its surrounding environment. Here, we see that RMP-VICES in these domains exhibits a sharper distribution about 0N (Fig. \ref{fig:training_collision_hists}).

%===============================================================================

\section{Conclusion}
\label{sec:conclusion}
% Tuning problem
% Tree-structure limits information passing between leaf node
We have presented RMP-VICES, a geometry-aware operational-space controller for contact-rich manipulation. Our formulation synthesizes learned impedance control in $\mathsf{SE(3)}$ with predefined collision and joint limit avoidance behaviors, yielding a safer framework for reinforcement learning of manipulation skills. 
Our evaluation shows reduced frequency of collision and joint limit violation on the majority of tasks studied, especially early in learning.
The main drawbacks of our proposed approach are the intensive tuning efforts required to weight leaf policies appropriately in the RMP-tree. Our framework also lacks Lyapunov or control barrier-certificates that guarantee stability or safety set invariance, respectively. 
Future work is warranted to investigate the design and tuning of provably safety behaviors and their interaction with policy optimization.

\section*{Acknowledgements}
We thank Roberto Martín-Martín for his incredibly detailed descriptions of the VICES experiments and openness for all of our inquiries about the work. We also thank Eric Rosen, Cameron Allen, and Akhil Bagaria for their help in forming our policy network structure.

This research was supported by NSF CAREER Award 1844960 to Konidaris, and by the ONR under the PERISCOPE MURI Contract N00014-17-1-2699. Disclosure: George Konidaris is the Chief Roboticist of Realtime Robotics, a robotics company that produces a specialized motion planning processor.

\bibliography{example.bib}{}
\bibliographystyle{IEEEtran}

\end{document}